\documentclass[10pt,twocolumn,letterpaper]{article}

\usepackage[pagenumbers]{iccv} 

\definecolor{iccvblue}{rgb}{0.21,0.49,0.74}
\usepackage[pagebackref,breaklinks,colorlinks,allcolors=iccvblue]{hyperref}
\usepackage{multirow}
\usepackage{soul}

\title{C2D-ISR: Optimizing Attention-based Image Super-resolution \\ from Continuous to Discrete Scales}

\author{
Yuxuan Jiang$^1$,
Chengxi Zeng$^1$,
Siyue Teng$^1$,
Fan Zhang$^1$,  \\
Xiaoqing Zhu$^2$,
Joel Sole$^2$,
and David Bull$^1$ \\
$^1$ \textit{Visual Information Laboratory, University of Bristol, Bristol, BS1 5DD, UK}\\
$^1$ \textit{\{yuxuan.jiang, simon.zeng, siyue.teng, fan.zhang, dave.bull\}@bristol.ac.uk}\\
$^2$ \textit{Netflix Inc., Los Gatos, CA, USA, 95032}\\
$^2$ \textit{\{xzhu, jsole\}@netflix.com}\\
}

\begin{document}
\maketitle
\begin{abstract}
In recent years, attention mechanisms have been exploited in single image super-resolution (SISR), achieving impressive reconstruction results. However, these advancements are still limited by the reliance on simple training strategies and network architectures designed for discrete up-sampling scales, which hinder the model's ability to effectively capture information across multiple scales. To address these limitations, we propose a novel framework, \textbf{C2D-ISR}, for optimizing attention-based image super-resolution models from both performance and complexity perspectives. Our approach is based on a two-stage training methodology and a hierarchical encoding mechanism. The new training methodology involves continuous-scale training for discrete scale models, enabling the learning of inter-scale correlations and multi-scale feature representation. In addition, we generalize the hierarchical encoding mechanism with existing attention-based network structures, which can achieve improved spatial feature fusion, cross-scale information aggregation, and more importantly, much faster inference. We have evaluated the C2D-ISR framework based on three efficient attention-based backbones, SwinIR-L, SRFormer-L and MambaIRv2-L, and demonstrated significant improvements over the other existing optimization framework, HiT, in terms of super-resolution performance (up to 0.2dB) and computational complexity reduction (up to 11\%). The source code will be made publicly available at \url{www.github.com}.
\end{abstract}

\section{Introduction}
\label{sec:intro}

\begin{figure}
    \centering
		  \includegraphics[width=1\linewidth]{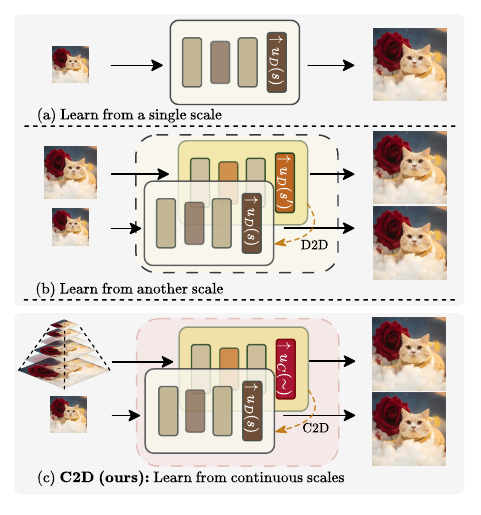}
	
    \caption{Different ISR training strategies: (a) the conventional training methodology \cite{liang2021swinir, guo2024mambair, dong2015image} only involving training models at a fixed scale; (b) the training method based on discrete multiple scales, which can relatively improve single scale ISR performance \cite{kim2016accurate, lim2017enhanced, lai2017deep}; (c) the proposed \textbf{C2D} training strategy, which employs a new implicit image function to learn inter-scale correlations from continuous ISR models - this strategy maintains low computational complexity and enhances the overall performance. $u_{C}$ denotes the continuous up-sampler, while $u_{D}$ represents the discrete up-sampler. $s$ and $s'$ are the up-sampling scales, and $\sim$ stands for continuous scales.}
    \label{fig:com}
\end{figure}

Image Super-Resolution (ISR) is one of the fundamental tasks in low-level computer vision that aims to reconstruct a high-resolution (HR) image from a corresponding low-resolution (LR) version. In the past decade, inspired by the advances in deep learning, numerous learning-based ISR methods have been developed for various applications spanning medical imaging \cite{georgescu2023multimodal, li2021review}, satellite remote sensing \cite{wang2022comprehensive, xiao2021satellite}, video streaming \cite{afonso2018video, zhang2021vistra2}, and surveillance \cite{zhang2010super, wang2020deep}. By leveraging data-driven feature extraction and high-capacity networks, these methods have outperformed conventional up-sampling methods based on classic signal processing theories. Among these learning-based ISR models, some \cite{liang2022vrt, zhu2023attention, wang2022uformer, liang2021swinir, conde2022swin2sr, chen2023hat} employ attention mechanisms, e.g. Vision Transformers (ViTs), to capture global dependencies, often resulting in improved reconstruction performance compared to convolutional neural network based approaches \cite{dong2015image, kim2016accurate, lim2017enhanced, zhang2018image}. However, a primary concern with existing ViT-based ISR methods is the poor trade-off between performance and complexity, in particular when the application demands real-time deployment.

To mitigate the increased computational complexity associated with attention-based ISR models, various lightweight ISR models have been proposed based on different parameter reduction strategies \cite{jiang2023compressing, zhang2021learning, yu2023dipnet, kim2016deeply, tai2017memnet}. When these methods are applied to complex ISR models, significant complexity reduction can be achieved, but only at the expense of lower reconstruction quality. Although advanced training approaches exist, such as knowledge distillation \cite{jiang2024mtkd, he2020fakd, fang2022cross}, which helps to improve lightweight model performance, these lightweight ISR models are still constrained by their limited network capacity.

In this context, inspired by the recent advances in continuous image super-resolution \cite{chen2021learning, lee2022local, jiang2024hiif} and hierarchical encoding mechanisms \cite{kwan2023hinerv, li2024asmr}, this paper proposes a novel framework, \textbf{C2D-ISR}, for optimizing lightweight attention-based ISR models for discrete up-sampling scales. C2D-ISR is based on a new two-stage training strategy, which first pre-trains the target model on a continuous scale, allowing the network to fully learn the mapping between multiple scales (as shown in \autoref{fig:com}). In the second stage, a more efficient and adaptable model for a single scale is then used to replace the pre-trained continuous INR up-sampler, which is fine-tuned on that fixed scale. This unique design not only inherits the detail capturing ability and multi-scale features gained from continuous-scale training, but also maintains the low complexity of the ISR model associated with the discrete-scale up-sampler. Moreover, to further trade off between performance and model complexity, we introduce a new multi-scale network design based on hierarchical encoding \cite{kwan2023hinerv} which can more effectively extract and fuse features across different levels, thereby balancing global semantics with local details. The primary contributions of this work are summarized as follows:

\begin{itemize}
    \item A new optimization framework is proposed for enhancing attention-based super-resolution networks. This, for the {first time}, enables a discrete scale ISR model to \textbf{learn from its continuous scale version}, resulting in consistently improved overall performance.
    
    \item A new hierarchical encoding based network structure, \textbf{Hi}erarchical \textbf{E}ncoding \textbf{T}ransformer (\textbf{HiET}) layer, has been designed to replace conventional attention-based networks. This captures information across multiple scales, resulting in reduced inference complexities and better performance. As far as we are aware, this is the \textbf{first time} that \textbf{hierarchical encoding} has been integrated with attention mechanisms for super-resolution.

    \item A \textbf{new U-Net architecture with adaptive window sizes} connecting different scale information has been devised to assemble HiET layers at various levels, in order to enhance multi-scale feature fusion. Unlike the traditional U-Net \cite{ronneberger2015u}, which relies on convolution and upsampling for resolution changes, our approach adjusts the feature resolution by modifying the window size in the HiET layer at different levels. This achieves consistent performance improvements with minimal computational overhead.
\end{itemize}

We have integrated the proposed C2D-ISR framework with three efficient attention-based backbones, SwinIR-L \cite{liang2021swinir}, SRFormer-L \cite{zhou2023srformer} and MambaIRv2-L \cite{guo2024mambairv2} and benchmarked it against HiT \cite{zhang2024hit}, the only comparable optimization framework in the literature. Experimental results demonstrate that C2D-ISR consistently outperforms HiT across different backbones, databases and SR tasks, resulting in optimized models with even lower complexities. 

\begin{figure*}[!t]
    \centering
		  \includegraphics[width=1\linewidth, trim=60 0 0 50]{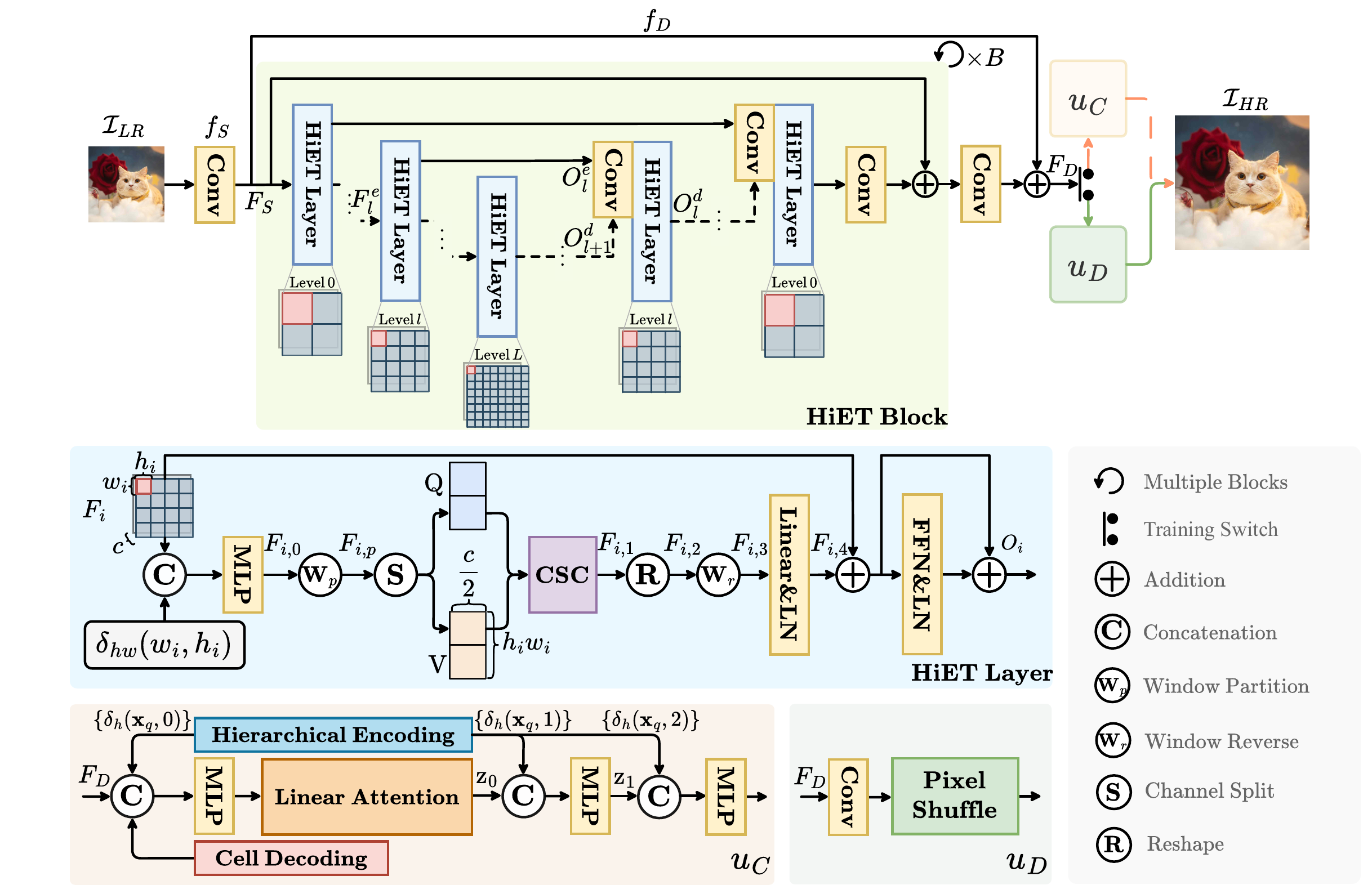}
    \caption{The architecture of the proposed C2D-ISR framework. (Top) the overall architecture and the design of the HiET block. The hyperparameter $B$ stands for the number of HiET blocks in the deep feature extractor $f_{D}$. (Middle) the design of the HiET layer. (Bottom) the up-sampler used for continuous-scale and discrete-scale training.}
    \label{fig:archi}
\end{figure*}

\section{Related Work}
\label{sec:RelatedWork}

\noindent\textbf{Image super-resolution (ISR).} Existing learning-based ISR methods are typically based on one of three popular architectures: convolutional neural networks (CNNs) \cite{dong2015image, kim2016accurate, lim2017enhanced, zhang2018image}, vision transformers (ViTs) \cite{liang2022vrt, wang2022uformer, liang2021swinir, conde2022swin2sr, chen2023hat}, and structured state-space models \cite{guo2024mambair, ren2024mambacsr, shi2024vmambair}. CNN-based models often perform feature extraction and residual learning to obtain spatial details in high-resolution content, while transformer-based models \cite{liang2021swinir, lu2022transformer, chen2023hat} leverage attention mechanisms to capture global dependencies, further refining SR performance. Additionally, structured state-space models build upon self-attention modules and introduce structured state-space sequences, which have been reported to outperform transformer-only architectures \cite{guo2024mambair, guo2024mambairv2}. Although these advanced network structures have achieved consistent performance improvements, the growing complexity of attention-based models also poses challenges for practical, in particular real-time, applications.

\vspace{5pt}

\noindent\textbf{Efficient image super-resolution.} To address the high complexity of attention-based ISR models, numerous lightweight architectures have been proposed that trade off efficiency and performance. Early methods focused on kernel weight reuse \cite{kim2016deeply, tai2017image} to reduce the number of parameters, but these had a limited impact on efficiency. Later approaches introduced compact architectures based on cascading residual blocks \cite{ahn2018fast}, information multi-distillation \cite{ahn2018fast}, lattice filtering \cite{luo2020latticenet} and structure pruning \cite{zhang2021learning, jiang2023compressing}. More recently, lightweight network structures have been designed including SwinIR-L \cite{liang2021swinir}, HNCT \cite{fang2022hybrid}, ESRT \cite{lu2022transformer}, ELAN \cite{zhang2022efficient}, and Omni-SR \cite{wang2023omni}. Moreover, HiT-SR \cite{zhang2024hit} proposes a spatial-channel correlation (SCC) method with linear computational complexity, while in \cite{park2025efficient}, attention sharing has been proposed to alleviate the efficiency bottleneck of a self-attention layer.

\vspace{5pt}

\noindent\textbf{Continuous super-resolution (CSR)} aims to generate high-resolution (HR) images for arbitrary scaling factors, offering greater flexibility than traditional ISR methods that are typically limited to fixed up-sampling scales. Most CSR methods involve implicit image functions, which model image details continuously. Notable CSR models include MetaSR \cite{hu2019meta}, LIIF \cite{chen2021learning}, LTE \cite{lee2022local} and HIIF \cite{jiang2024hiif}, all of which can achieve improved performance compared to their corresponding discrete scale versions. However, this characteristic has not been exploited when discrete ISR models are designed and optimized.

\vspace{5pt}

\noindent\textbf{Hierarchical encoding} \cite{kwan2023hinerv} is a new technique used in deep learning to effectively encode information at multiple levels of abstraction. It enables the resulting models to progressively extract and refine features at different resolutions, enhancing feature representation and improving performance for various tasks \cite{li2024asmr, kwan2024nvrc}. Unlike conventional single-level grid-based encoding that relies on global coordinates for position computation, hierarchical encoding leverages local coordinates to encode relative positional information. Due to the much smaller value ranges with local coordinates, the required feature grid is more compact - this can significantly reduce complexity overhead.

\section{Methods}

A typical attention-based image super-resolution framework 
consists of three key components: a shallow feature extractor, $f_{S}$, a deep feature extractor, $f_{D}$, and an up-sampler, $u$. Given an input low-resolution (LR) image, $\mathcal{I}_{LR} \in \mathbb{R}^{H \times W \times 3}$, the shallow feature extractor $f_{S}$ employs a convolution layer to project $\mathcal{I}_{LR}$ into a feature space $F_{S} = f_{S}(\mathcal{I}_{LR}) \in \mathbb{R}^{H \times W \times C}$, where $C$ is the number of channels. The deep feature extractor $f_{D}$ then processes $F_{S}$ through a series of transformer-based blocks, followed by another convolution layer, $F_{D} = f_{D}(F_{S}) \in \mathbb{R}^{H \times W \times C}$. Here, each transformer block contains multiple transformer-based layers, each of which contains a self-attention network (SA), a feedforward network (FFN), and layer normalization (LN). Finally, the up-sampler reconstructs the high-resolution (HR) image $\mathcal{I}_{HR} \in \mathbb{R}^{sH \times sW \times 3}$ from $F_{D}$, where $s$ denotes the up-scaling factor. 

In this work, our aim is to optimize the training strategy and the design of the attention network for the aforementioned framework. We first designed a two-stage training strategy, which optimizes the proposed ISR model by learning from its continuous scale counterpart while maintaining its relatively low computational complexity. To further improve the performance of the proposed model, we also designed a novel Hierarchical Encoding Transformer (HiET) layer that is integrated into HiET blocks based on a modified U-Net style architecture. This provides the backbone of the deep feature extractor $f_D$ in this work,  enhancing multi-scale information fusion and improving feature aggregation across different scales. The HiET layer leverages a hierarchical encoding mechanism to capture structural information efficiently while maintaining low computational complexity. These design details are described below. 

\subsection{HiET Block}

Although HiT \cite{zhang2024hit} adopts a variable window size mechanism to capture multi-scale information, the employed sequential arrangement only allows each layer to aggregate information from the previous one, which limits the learning of rich multi-scale features. In this work, we address this issue by incorporating a modified U-Net architecture with skip connections between paired layers and adaptive window sizes to enhance multi-scale information fusion.

Specifically, as illustrated in \autoref{fig:archi}. (top), each Hierarchical Encoding Transformer (HiET) block contains an encoder with $L+1$ HiET layers and a decoder with $L$ HiET layers. In the following description, we use superscripts $^e$ and $^d$ to differentiate their locations (in the encoder or decoder). In the encoder, the window size of the attention decreases from lower to higher layers. For each encoder layer (with index $l$), the input feature $F^{e}_l$ is the output of the lower layer $O^{e}_{l-1}$. On the contrary, at the decoder, the window size of the attention increases at lower levels. For each decoder layer (with index $l$), its input $F^{d}_l$ is formed by concatenating the output from the $(l+1)$-th decoder layer $O^{d}_{l+1}$, and that from $l$-th encoder layer before passing the concatenated features through a 1$\times$1 convolution layer:
\begin{equation}
    F^{d}_l = \mathrm{Conv}([O^{d}_{l+1}, O^{e}_{l}]).
\end{equation}

\subsection{HiET Layer}
Inspired by the hierarchical encoding mechanism proposed in \cite{kwan2023hinerv, jiang2024hiif}, we designed a new attention (HiET) layer, which incorporates such hierarchical encoding with multi-scale features to better capture complex structural information and spatial dependencies. Specifically, as shown in \autoref{fig:archi}, in each HiET layer (within the encoder or decoder, indexed by $i$), given the input feature $F_i \in \mathbb{R}^{H \times W \times C}$, with window size ($w_i$, $h_i$), the embedded hierarchical encoding $\delta_{hw}$ at feature index ($u$,$v$) is defined as:
\begin{equation}
    \delta_{hw} (w_i, h_i) = (\lfloor \frac{u}{w_i}\times2\rfloor \bmod 2, \lfloor \frac{v}{h_i}\times2\rfloor \bmod 2),
\end{equation}
where $u = 0, 1, \dots, W-1$, $v = 0, 1, \dots, H-1.$ It is concatenated with the input feature $F_i$ and passed through a multilayer perceptron (MLP), resulting in $F_{i,0}$:
\begin{equation}
    F_{i,0} = \mathrm{MLP}([F_{i}, \{\delta_{hw} (w_i, h_i)\}]),
\end{equation}
where $\{\cdot\}$ stands for the set of all points, i.e. $\{\delta_{hw} (w_i, h_i)\} \in \mathbb{R}^{H \times W \times 2}$. $F_{i,0}$ is then partitioned into non-overlapped subblocks according to the corresponding window size ($w_i$, $h_i$) and each of them is reshaped as $F_{i,p} \in \mathbb{R}^{h_{i}w_{i} \times C}$, where subblock index is omitted for simplicity. A channel-splitting operation is performed after reshaping to obtain two sets of features, $Q$ and $V$:
\begin{equation}
    [Q, V] = \mathrm{split}(F_{i,p}),
\end{equation}
where $Q, V \in \mathbb{R}^{h_{i}w_{i} \times \frac{C}{2}}$. We then apply Channel Self-Correlation (CSC) \cite{zhang2024hit} to leverage different window sizes and effectively utilize rich multi-scale information for performance enhancement while maintaining low complexity,
\begin{equation}
    F_{i,1} = (\frac{Q^{T}V}{h_{i}w_{i}}V^{T})^T .
\end{equation}
Here $(\cdot)^T$ represents matrix transpose, and $F_{i,1} \in \mathbb{R}^{h_{i}w_{i} \times \frac{C}{2}}$. $F_{i,1}$ is then reshaped into $F_{i,2} \in \mathbb{R}^{h_{i} \times w_{i} \times \frac{C}{2}}$, and reversed to match the spatial resolution of $F_i$, resulting in $F_{i,3} \in \mathbb{R}^{H \times W \times \frac{C}{2}}$. To restore the channel dimension, we apply a linear transformation followed by layer normalization, resulting in $F_{i,4} \in \mathbb{R}^{H \times W \times C}$:
\begin{equation}
    F_{i,4} = \mathrm{LN}(\mathrm{Linear}(F_{i,3})).
\end{equation}
Finally, through a skip connection and feedforward network (FFN), the output of this layer is obtained, denoted as $O_{i}$.

\subsection{Training Methodology: C2D}

As mentioned above, all existing ISR optimization frameworks \cite{liang2021swinir,zhang2024hit,guo2024mambair,zhang2022efficient, zhou2023srformer} adopt simple training strategies based on a fixed up-sampling factor. It is noted that continuous super-resolution methods can deal with arbitrary up-sampling ratios with a single model and exhibit improved performance over discrete ISR approaches at their pre-trained scales. However, this characteristic has not been exploited during the training of discrete ISR models. To this end, we designed a novel two-stage training methodology to boost the training performance of our optimization framework, which consists of two stages: i) Continuous-Scale Pre-Training; ii) Discrete-Scale Fine-Tuning.

\vspace{5pt}

\noindent\textbf{Stage 1: Continuous-Scale Pre-Training. } We first optimize a continuous-scale model designed to exploit inter-scale correlations. By learning how to up-sample to arbitrary scales, the model is expected to learn the correlation between multiple scales and better recover high-frequency details. Inspired by  \cite{jiang2024hiif}, we designed a lightweight implicit image function, HIIF-L, to achieve continuous-scale up-sampling ($u_C$ in~\autoref{fig:archi}). HIIF-L is associated with a relatively small model size (compared to the original HIIF \cite{jiang2024hiif}), only consisting of three MLPs and a linear attention module. This will enable sufficient optimization of the HiET blocks rather than relying on the lightweight HIIF-L to obtain satisfactory super-resolution performance.

Specifically, the input of the first MLP is the concatenation of the deep feature $F_{D}$ obtained from $f_D$ and $cell = [\frac{2}{sH}, \frac{2}{sW}]$ that represents cell decoding and hierarchical coordinates, $s$ stands for the up-sampling factor. 
Following \cite{jiang2024hiif}, the hierarchical coordinate $\delta_{h} (\mathbf{x}_{q}, j)$ is derived from the local coordinate $\mathbf{x}_{q} = (x_{local}, y_{local})$:
\begin{equation}
    \delta_{h} (\mathbf{x}_{q}, j) = \lfloor \mathbf{x}_{q} \times 2^{j + 1}\rfloor \bmod 2,
\end{equation}
where $j$ indexes the current level of the hierarchy. Intuitively, this allows HIIF-L to handle multi-scale structures by encoding fine-grained positional information. The resulting features then undergo a linear attention mechanism that modulates global-local context:
\begin{align}
 \mathbf{z_{0}} &= \mathrm{MHA}(\mathrm{MLP}([F_{D}, \{\delta_{h} (\mathbf{x}_{q}, 0)\}, cell])),
\end{align}
where $\mathbf{z_{0}} \in \mathbb{R}^{sH \times sW \times dim}$ denotes the feature output of the linear attention. In this work, we set $dim$ equal to $C$. The refined features are then fed into the second MLP, which predicts the final high-resolution outputs $I_{HR} \in \mathbb{R}^{sH \times sW \times 3}$:
\begin{align}
 \mathbf{z_{1}} &= \mathrm{MLP}([\mathbf{z_{0}}, \{\delta_{h} (\mathbf{x}_{q}, 1)\}]).
\end{align}
This design enables HIIF-L to seamlessly adapt to different scale factors, preserving both global structures and high-frequency details.

\vspace{5pt}

\noindent\textbf{Stage 2: Discrete-Scale Fine-Tuning.} In the second stage, we replaced HIIF-L with a commonly used sub-pixel convolution ($u_D$ in \autoref{fig:archi}) and trained the resulting ISR model for three fixed scaling factors ($\times$2, $\times$3, and $\times4$). We first initialize the weights of $f_S$ and $f_D$ with the model parameters obtained in Stage 1, and then fine-tune them alongside the discrete up-sampler ($u_D$). These fine-tuned weights are ultimately used for model inference at the target scales. 
\begin{equation}
    \mathcal{I}_{HR} = \mathrm{Pixelshuffle}(\mathrm{Conv}(F_{D})).
\end{equation}

\begin{table*}[!ht]
\centering

\resizebox{\linewidth}{!}{\begin{tabular}{l|c|cc|cc|cc|cc|cc|cc}
\toprule
\multicolumn{2}{c|}{} & \multicolumn{2}{c|}{Complexity} & \multicolumn{2}{c|}{Set5} & \multicolumn{2}{c|}{Set14} & \multicolumn{2}{c|}{BSD100} & \multicolumn{2}{c|}{Urban100} & \multicolumn{2}{c}{Manga109} \\ \midrule
Method & Scale & \#Para.$\downarrow$ & FLOPs$\downarrow$ & PSNR$\uparrow$ & SSIM$\uparrow$ & PSNR & SSIM & PSNR & SSIM & PSNR & SSIM & PSNR & SSIM \\ \midrule
 SwinIR-L\cite{liang2021swinir} & \multirow{3}{*}{$\times$2} & 910K &244.4G & 38.14  & 0.9611 & 33.86  & 0.9206 & 32.31  & 0.9012 & 32.76  & 0.9340 & 39.12  & 0.9783 \\
SwinIR-HiT\cite{zhang2024hit} &  & 772K &209.9G & 38.22  & \textcolor{blue}{0.9613} & 33.91  & \textcolor{blue}{0.9213} & 32.35  & 0.9019 & 33.02  & 0.9365 & 39.38  & 0.9782 \\
\textbf{SwinIR-C2D (ours)} & & \textcolor{blue}{755K} &\textcolor{blue}{201.2G} & \textcolor{blue}{38.27}  & 0.9612 & \textcolor{blue}{33.98}  & 0.9211 & \textcolor{blue}{32.39}  & \textcolor{blue}{0.9020} & \textcolor{blue}{33.14}  & \textcolor{blue}{0.9370} & \textcolor{blue}{39.43}  & \textcolor{blue}{0.9784}  \\ \midrule 
SwinIR-L\cite{liang2021swinir} & \multirow{3}{*}{$\times$3} & 918K &110.8G & 34.62  & 0.9289 & 30.54  & 0.8463 & 29.20  & 0.8082 & 28.66  & 0.8624 & 33.98  & 0.9478 \\
SwinIR-HiT\cite{zhang2024hit} &  & 780K &94.2G & 34.72  & 0.9298 & 30.62  & \textcolor{blue}{0.8474} & 29.27  & 0.8101 & 28.93  & 0.8673 & 34.40  & \textcolor{blue}{0.9496} \\
\textbf{SwinIR-C2D (ours)} & & \textcolor{blue}{763K} &\textcolor{blue}{90.6G} & \textcolor{blue}{34.81}  & \textcolor{blue}{0.9299} & \textcolor{blue}{30.65}  & 0.8473 & \textcolor{blue}{29.33}  & \textcolor{blue}{0.8099} & \textcolor{blue}{29.04}  & \textcolor{blue}{0.8687} & \textcolor{blue}{34.46}  & 0.9494 \\ \midrule 
SwinIR-L\cite{liang2021swinir} & \multirow{3}{*}{$\times$4} & 
930K &63.6G & 32.44  & 0.8976 & 28.77  & 0.7858 & 27.69  & 0.7406 & 26.47  & 0.7980 & 30.92  & 0.9151\\
SwinIR-HiT\cite{zhang2024hit} &  & 792K &53.8G & 32.51  & 0.8991 & 28.84  & 0.7873 & 27.73  & 0.7424 & 26.71  & 0.8045 & 31.23  & 0.9176 \\
\textbf{SwinIR-C2D (ours)} & & \textcolor{blue}{775K} &\textcolor{blue}{52.7G} & \textcolor{blue}{32.61}  & \textcolor{blue}{0.8997} & \textcolor{blue}{28.89}  & \textcolor{blue}{0.7875} & \textcolor{blue}{27.77}  & \textcolor{blue}{0.7426} & \textcolor{blue}{26.87}  & \textcolor{blue}{0.8063} & \textcolor{blue}{31.33}  & \textcolor{blue}{0.9181} \\ \midrule\midrule
SRFormer-L\cite{zhou2023srformer} & \multirow{3}{*}{$\times$2}  & 853K &236.2G & 38.23  & 0.9613 & 33.94  & 0.9209 & 32.36  & 0.9019 & 32.91  & 0.9353 & 39.28  & 0.9785 \\
SRFormer-HiT\cite{zhang2024hit} &  & 847K &226.5G & 38.26  & 0.9615 & 34.01  & 0.9214 & 32.37  & 0.9023 & 33.13  & 0.9372 & 39.47  & 0.9787 \\
\textbf{SRFormer-C2D (ours)} &  & \textcolor{blue}{830K} &\textcolor{blue}{218.4G} & \textcolor{blue}{38.33}  & \textcolor{blue}{0.9619} & \textcolor{blue}{34.18}  & \textcolor{blue}{0.9225} & \textcolor{blue}{32.45}  & \textcolor{blue}{0.9028} & \textcolor{blue}{33.28}  & \textcolor{blue}{0.9382} & \textcolor{blue}{39.53}  & \textcolor{blue}{0.9789}  \\ \midrule
SRFormer-L\cite{zhou2023srformer} & \multirow{3}{*}{$\times$3}  & 861K &104.8G & 34.67  & 0.9296 & 30.57  & 0.8469 & 29.26  & 0.8099 & 28.81  & 0.8655 & 34.19  & 0.9489 \\
SRFormer-HiT\cite{zhang2024hit} &  & 855K &101.6G & 34.75  & 0.9300 & 30.61  & 0.8475 & 29.29  & 0.8106 & 28.99  & 0.8687 & 34.53  & \textcolor{blue}{0.9502} \\
\textbf{SRFormer-C2D (ours)} &  & \textcolor{blue}{838K} &\textcolor{blue}{98.3G} & \textcolor{blue}{34.84}  & \textcolor{blue}{0.9302} & \textcolor{blue}{30.71}  & \textcolor{blue}{0.8488} & \textcolor{blue}{29.34}  & \textcolor{blue}{0.8108} & \textcolor{blue}{29.17}  & \textcolor{blue}{0.8707} & \textcolor{blue}{34.57}  & 0.9499  \\ \midrule
SRFormer-L\cite{zhou2023srformer} & \multirow{3}{*}{$\times$4}  & 873K &62.8G & 32.51  & 0.8988 & 28.82  & 0.7872 & 27.73  & 0.7422 & 26.67  & 0.8032 & 31.17  & 0.9165 \\
SRFormer-HiT\cite{zhang2024hit} &  & 866K &58.0G & 32.55  & 0.8999 & 28.87  & 0.7880 & 27.75  & 0.7432 & 26.80  & 0.8069 & 31.26  & 0.9171 \\
\textbf{SRFormer-C2D (ours)} &  & \textcolor{blue}{849K} &\textcolor{blue}{57.0G} & \textcolor{blue}{32.67}  & \textcolor{blue}{0.9002} & \textcolor{blue}{28.91}  & \textcolor{blue}{0.7883} & \textcolor{blue}{27.80}  & \textcolor{blue}{0.7439} & \textcolor{blue}{26.96}  & \textcolor{blue}{0.8104} & \textcolor{blue}{31.45}  & \textcolor{blue}{0.9180}  \\ \midrule\midrule
MambaIRv2-L\cite{guo2024mambairv2} & \multirow{3}{*}{$\times$2} & 774K &286.3G & 38.26  & 0.9615 & 34.09  & 0.9221 & 32.36  & 0.9019 & 33.26  & 0.9378 & 39.35  & 0.9785 \\
MambaIRv2-HiT$^\dagger$\cite{zhang2024hit} &  & 638K &164.5G & 38.28  & 0.9615 & 34.11  & 0.9225 & 32.40  & \textcolor{blue}{0.9022} & 33.30  & 0.9387 & 39.38  & 0.9788  \\
\textbf{MambaIRv2-C2D (ours)} &  & \textcolor{blue}{631K} &\textcolor{blue}{152.9G} & \textcolor{blue}{38.31}  & \textcolor{blue}{0.9618} & \textcolor{blue}{34.16}  & \textcolor{blue}{0.9228} & \textcolor{blue}{32.45}  & 0.9021 & \textcolor{blue}{33.40}  & \textcolor{blue}{0.9396} & \textcolor{blue}{39.43}  & \textcolor{blue}{0.9792}  \\ \midrule
MambaIRv2-L\cite{guo2024mambairv2} & \multirow{3}{*}{$\times$3} & 781K &126.7G & 34.71  & 0.9298 & 30.68  & 0.8483 & 29.26  & 0.8098 & 29.01  & 0.8689 & 34.41  & 0.9497 \\
MambaIRv2-HiT$^\dagger$\cite{zhang2024hit} &  & 645K &77.5G & 34.75  & 0.9299 & 30.70  & 0.8480 & 29.27  & 0.8099 & 29.07  & 0.8693 & 34.50  & 0.9500  \\
\textbf{MambaIRv2-C2D (ours)}  &  & \textcolor{blue}{638K} & \textcolor{blue}{69.0G} & \textcolor{blue}{34.78}  & \textcolor{blue}{0.9303} & \textcolor{blue}{30.78}  & \textcolor{blue}{0.8488} & \textcolor{blue}{29.36}  & \textcolor{blue}{0.8105} & \textcolor{blue}{29.15}  & \textcolor{blue}{0.8701} & \textcolor{blue}{34.56}  & \textcolor{blue}{0.9503} \\ \midrule
MambaIRv2-L\cite{guo2024mambairv2} & \multirow{3}{*}{$\times$4} & 790K &75.6G & 32.51  & 0.8992 & 28.84  & 0.7878 & 27.75  & 0.7426 & 26.82  & 0.8079 & 31.24  & 0.9182 \\
MambaIRv2-HiT$^\dagger$\cite{zhang2024hit} &  & 653K &42.0G & 32.50  & 0.8990 & 28.84  & 0.7877 & 27.78  & 0.7427 & 26.84  & 0.8083 & 31.27  & 0.9186  \\
\textbf{MambaIRv2-C2D (ours)}  &  & \textcolor{blue}{647K} &\textcolor{blue}{40.0G} & \textcolor{blue}{32.59}  & \textcolor{blue}{0.8996} & \textcolor{blue}{28.89}  & \textcolor{blue}{0.7883} & \textcolor{blue}{27.86}  & \textcolor{blue}{0.7432} & \textcolor{blue}{26.96}  & \textcolor{blue}{0.8097} & \textcolor{blue}{31.37}  & \textcolor{blue}{0.9195} \\ \bottomrule
\end{tabular}}
\caption{Quantitative comparison with state-of-the-art SR methods on the Set5 \cite{bevilacqua2012low}, Set14 \cite{zeyde2012single}, BSD100 \cite{martin2001database}, and Urban100 \cite{huang2015single} and Manga109 \cite{matsui2017sketch}. The output size is set to 720 $\times$ 1280 for all scales to compute FLOPs. The symbol $^\dagger$ indicates that the corresponding benchmark model was implemented by us following its original literature. The best result is colored in \textcolor{blue}{blue}.}
\label{tbl:Results}
\end{table*}

\section{Experiment Configuration}

\noindent{\textbf{Implementation details.}} We adopt L1 as the loss function during the training and use the Adam \cite{kingma2014adam} optimizer. In the first training stage (at the continuous scale), the maximum and minimum learning rates are set to $4 \times 10^{-4}$ and $1 \times 10^{-6}$, respectively, while in the second stage (at the discrete scale), their values are $1 \times 10^{-5}$ and $1 \times 10^{-6}$ instead. Models are trained for 1000 epochs with a batch size of 16 in each stage: 700 epochs for the first stage and 300 epochs for the second stage. Following a 50-epoch warm-up phase, the learning rate is decayed according to a cosine annealing schedule \cite{loshchilov2016sgdr}. The training and inference of all (proposed and benchmark) models were based on a single NVIDIA RTX 4090 graphics card. 

\vspace{5pt}

\noindent{\textbf{Baseline models.}} Following \cite{zhang2024hit}, we applied our C2D-ISR optimization framework to three latest lightweight ISR methods, SwinIR-L \cite{liang2021swinir}, SRFormer-L \cite{zhou2023srformer} and MambaIRv2-L \cite{guo2024mambairv2}. Compared to \cite{zhang2024hit}, we replaced one of the anchor models, SwinIR-NG, with a more recent model, MambaIRv2-L \cite{guo2024mambairv2}. All C2D-ISR optimized models retain the same hyperparameter settings as their corresponding original versions. Specifically, for SwinIR-L \cite{liang2021swinir} and SRFormer-L \cite{zhou2023srformer}, we set window sizes to [64, 32, 8, 8, 32, 64] for the six layers in each block respectively; for MambaIRv2-L \cite{guo2024mambairv2}, window sizes are set to [64, 32, 8, 32, 64] for the five layers in each block. All other settings, such as the number of blocks $B=4$, the number of channels $C$, and FFN design, remain unchanged as the original models. For each anchor model, we compare our approach with HiT \cite{zhang2024hit}, the only existing comparable method in the literature, and the pre-trained, unoptimized version. To offer additional benchmarks, we also provide results for other efficient ISR approaches, including HiT-SNG \cite{zhang2024hit}, SwinIR-NG \cite{choi2023n}, EDSR-B \cite{lim2017enhanced}, CARN \cite{ahn2018fast}, IMDN \cite{hui2019lightweight}, LatticeNet \cite{luo2020latticenet}, RFDN-L \cite{liu2020residual}, FMEN \cite{du2022fast}, GASSL-B \cite{wang2023global}, HNCT \cite{fang2022hybrid}, ELAN-L \cite{zhang2022efficient} and Omni-SR \cite{wang2023omni} in this paper\footnote{The results of these additional benchmarks are summarized in the \textit{Supplementary} due to the limited space, but are used to plot \autoref{fig:curves}.}.

\vspace{5pt}

\noindent{\textbf{Datasets.}} Following the previous works \cite{lim2017enhanced, chen2021learning, zhang2024hit}, we use the DIV2K training dataset \cite{agustsson2017ntire} from the NTIRE 2017 Challenge \cite{timofte2017ntire} for network optimization, which consists of 800 images in 2K resolution. In Stage 1, we utilize the training methodology described in \cite{chen2021learning, jiang2024hiif}, and sample $B_s$ random scaling factors $s_{1\cdots B_s}$ from a uniform distribution $\mathcal{U}(1,4)$, i.e. in-scale, where $B_s$ is the batch size. We also use the same scale factor for height and width, i.e. $s_x = s_y = s$, to crop ${64s \times 64s}$ patches from original images and generate their 64$\times$64 down-sampled counterparts based on bicubic resizing \cite{paszke2019pytorch}. In Stage 2, we fix the $s$ to 2, 3 and 4 for different discrete scales, $\times 2$, $\times 3$ and $\times 4$ respectively. Data augmentation strategies, such as random rotations and horizontal flips, are also applied during model training.

\vspace{5pt}
\noindent{\textbf{Evaluation Metrics.}} Five commonly used benchmark datasets: Set5 \cite{bevilacqua2012low}, Set14 \cite{zeyde2012single}, BSD100 \cite{martin2001database}, Urban100 \cite{huang2015single} and Manga109 \cite{matsui2017sketch}, are employed for model evaluation. Here, the LR images are obtained from their HR counterparts through bicubic degradation. We conduct comparisons under three up-scaling factors: $\times$2, $\times$3, and $\times$4, and assess the SR performance using PSNR and SSIM \cite{wang2004image}.

\begin{figure*}[!t]
    \centering
	\begin{minipage}{0.27\linewidth}
		  \centering
            \setlength{\abovecaptionskip}{0.cm}
		  \includegraphics[width=1\linewidth]{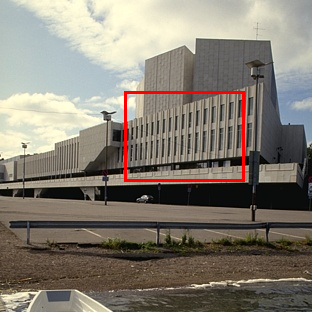}
            \caption*{\footnotesize 78004 (BSD100$, \times$4)}
	  \end{minipage}
      \begin{minipage}{0.7\textwidth}
        \centering
		\begin{minipage}{0.23\linewidth}
		  \centering
            \setlength{\abovecaptionskip}{0.cm}
		  \includegraphics[width=1\linewidth]{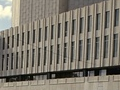}
            \caption*{\footnotesize HR }
	  \end{minipage}
        \begin{minipage}{0.23\linewidth}
		  \centering
            \setlength{\abovecaptionskip}{0.cm}
		  \includegraphics[width=1\linewidth]{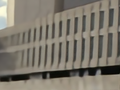}
            \caption*{\footnotesize MambaIRv2-L\cite{guo2024mambairv2} }
	  \end{minipage}
	  \begin{minipage}{0.23\linewidth}
		  \centering
            \setlength{\abovecaptionskip}{0.cm}
		  \includegraphics[width=1\linewidth]{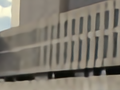}
            \caption*{\footnotesize MambaIRv2-HiT \cite{zhang2024hit}}
	  \end{minipage}
      	  \begin{minipage}{0.23\linewidth}
		  \centering
            \setlength{\abovecaptionskip}{0.cm}
		  \includegraphics[width=1\linewidth]{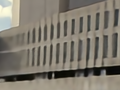}
            \caption*{\footnotesize MambaIRv2-C2D (ours)}
	  \end{minipage}

        \begin{minipage}{0.23\linewidth}
		  \centering
            \setlength{\abovecaptionskip}{0.cm}
		  \includegraphics[width=1\linewidth]{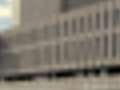}
            \caption*{\footnotesize Bicubic }
	  \end{minipage}
        \begin{minipage}{0.23\linewidth}
		  \centering
            \setlength{\abovecaptionskip}{0.cm}
		  \includegraphics[width=1\linewidth]{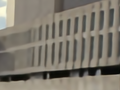}
            \caption*{\footnotesize SRFormer-L\cite{zhou2023srformer} }
	  \end{minipage}
	  \begin{minipage}{0.23\linewidth}
		  \centering
            \setlength{\abovecaptionskip}{0.cm}
		  \includegraphics[width=1\linewidth]{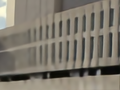}
            \caption*{\footnotesize SRFormer-HiT\cite{zhang2024hit}}
	  \end{minipage}
      	  \begin{minipage}{0.23\linewidth}
		  \centering
            \setlength{\abovecaptionskip}{0.cm}
		  \includegraphics[width=1\linewidth]{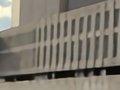}
            \caption*{\footnotesize SRFormer-C2D (ours)}
	  \end{minipage}
      \end{minipage}

 	\begin{minipage}{0.27\linewidth}
		  \centering
            \setlength{\abovecaptionskip}{0.cm}
		  \includegraphics[width=\linewidth]{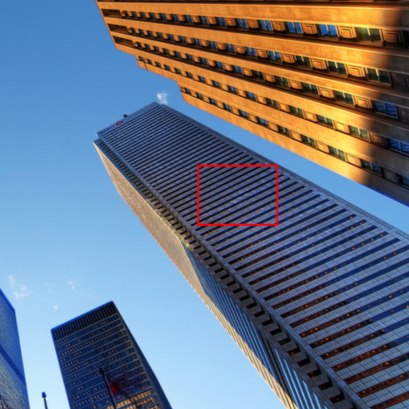}
            \caption*{\footnotesize img\_012 (Urban100$, \times$4)}
	  \end{minipage}
      \begin{minipage}{0.7\textwidth}
        \centering
		\begin{minipage}{0.23\linewidth}
		  \centering
            \setlength{\abovecaptionskip}{0.cm}
		  \includegraphics[width=1\linewidth]{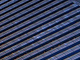}
            \caption*{\footnotesize HR }
	  \end{minipage}
        \begin{minipage}{0.23\linewidth}
		  \centering
            \setlength{\abovecaptionskip}{0.cm}
		  \includegraphics[width=1\linewidth]{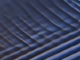}
            \caption*{\footnotesize SwinIR-L\cite{liang2021swinir} }
	  \end{minipage}
	  \begin{minipage}{0.23\linewidth}
		  \centering
            \setlength{\abovecaptionskip}{0.cm}
		  \includegraphics[width=1\linewidth]{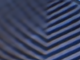}
            \caption*{\footnotesize SwinIR-HiT \cite{zhang2024hit}}
	  \end{minipage}
      	  \begin{minipage}{0.23\linewidth}
		  \centering
            \setlength{\abovecaptionskip}{0.cm}
		  \includegraphics[width=1\linewidth]{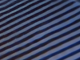}
            \caption*{\footnotesize SwinIR-C2D (ours)}
	  \end{minipage}

        \begin{minipage}{0.23\linewidth}
		  \centering
            \setlength{\abovecaptionskip}{0.cm}
		  \includegraphics[width=1\linewidth]{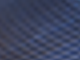}
            \caption*{\footnotesize Bicubic }
	  \end{minipage}
        \begin{minipage}{0.23\linewidth}
		  \centering
            \setlength{\abovecaptionskip}{0.cm}
		  \includegraphics[width=1\linewidth]{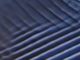}
            \caption*{\footnotesize SRFormer-L\cite{zhou2023srformer} }
	  \end{minipage}
	  \begin{minipage}{0.23\linewidth}
		  \centering
            \setlength{\abovecaptionskip}{0.cm}
		  \includegraphics[width=1\linewidth]{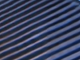}
            \caption*{\footnotesize SRFormer-HiT\cite{zhang2024hit}}
	  \end{minipage}
      	  \begin{minipage}{0.23\linewidth}
		  \centering
            \setlength{\abovecaptionskip}{0.cm}
		  \includegraphics[width=1\linewidth]{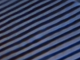}
            \caption*{\footnotesize SRFormer-C2D (ours)}
	  \end{minipage}
      \end{minipage}

      \begin{minipage}{0.27\linewidth}
		  \centering
            \setlength{\abovecaptionskip}{0.cm}
		  \includegraphics[width=\linewidth]{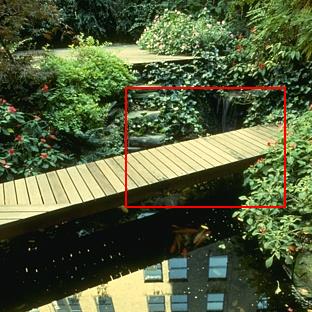}
            \caption*{\footnotesize 148026 (BSD100$, \times$4)}
	  \end{minipage}
      \begin{minipage}{0.7\textwidth}
        \centering
        \begin{minipage}{0.23\linewidth}
		  \centering
            \setlength{\abovecaptionskip}{0.cm}
		  \includegraphics[width=1\linewidth]{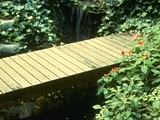}
            \caption*{\footnotesize LR }
	  \end{minipage}
        \begin{minipage}{0.23\linewidth}
		  \centering
            \setlength{\abovecaptionskip}{0.cm}
		  \includegraphics[width=1\linewidth]{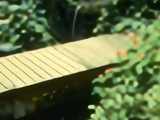}
            \caption*{\footnotesize MambaIRv2-L\cite{guo2024mambairv2} }
	  \end{minipage}
	  \begin{minipage}{0.23\linewidth}
		  \centering
            \setlength{\abovecaptionskip}{0.cm}
		  \includegraphics[width=1\linewidth]{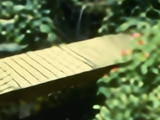}
            \caption*{\footnotesize MambaIRv2-HiT\cite{zhang2024hit}}
	  \end{minipage}
      	  \begin{minipage}{0.23\linewidth}
		  \centering
            \setlength{\abovecaptionskip}{0.cm}
		  \includegraphics[width=1\linewidth]{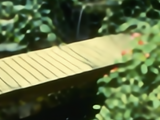}
            \caption*{\footnotesize MambaIRv2-C2D (ours)}
	  \end{minipage}

      \begin{minipage}{0.23\linewidth}
		  \centering
            \setlength{\abovecaptionskip}{0.cm}
		  \includegraphics[width=1\linewidth]{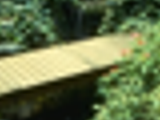}
            \caption*{\footnotesize Bicubic }
	  \end{minipage}
        \begin{minipage}{0.23\linewidth}
		  \centering
            \setlength{\abovecaptionskip}{0.cm}
		  \includegraphics[width=1\linewidth]{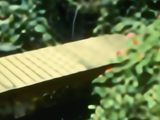}
            \caption*{\footnotesize SwinIR-L\cite{liang2021swinir} }
	  \end{minipage}
	  \begin{minipage}{0.23\linewidth}
		  \centering
            \setlength{\abovecaptionskip}{0.cm}
		  \includegraphics[width=1\linewidth]{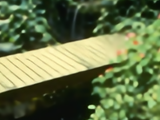}
            \caption*{\footnotesize SwinIR-HiT\cite{zhang2024hit}}
	  \end{minipage}
      	  \begin{minipage}{0.23\linewidth}
		  \centering
            \setlength{\abovecaptionskip}{0.cm}
		  \includegraphics[width=1\linewidth]{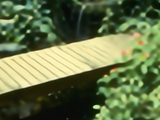}
            \caption*{\footnotesize SwinIR-C2D (ours)}
            \end{minipage}
      \end{minipage}
	
    \caption{Qualitative comparisons between the proposed C2D-SR framework and HiT \cite{zhang2024hit}.}
    \label{fig:Visualcomparisons}
\end{figure*}

\section{Results and Discussion}

\subsection{Overall Performance}

\begin{figure}
    \centering
		  \includegraphics[width=1\linewidth]{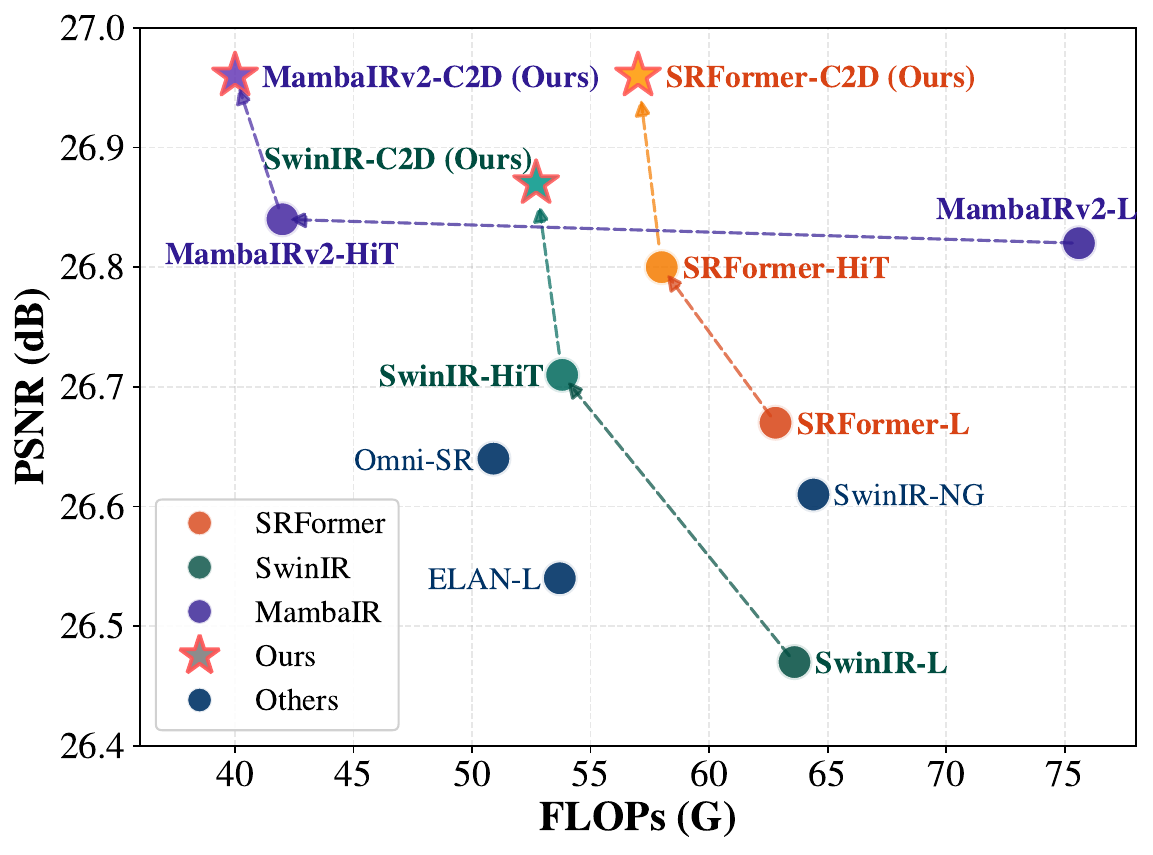}
	
    \caption{Complexity-performance trade-off visualization for selected ISR methods. The results are based on the Urban100 dataset and $\times$4 task.}
    \label{fig:curves}
\end{figure}

\noindent\textbf{Quantitative results.} \autoref{tbl:Results} summarizes the quantitative results, comparing the proposed C2D-ISR framework to HiT and the anchor lightweight model for three backbones, SwinIR-L \cite{liang2021swinir}, SRFormer-L \cite{zhou2023srformer} and MambaIRv2-L \cite{guo2024mambairv2}, in terms of PSNR, SSIM and complexity (model sizes and FLOPs). It can be observed that the proposed C2D-ISR consistently outperforms HiT (and the lightweight versions) across all backbones, databases, quality metrics and different up-sampling tasks ($\times$2, $\times$3 and $\times$4), with PSNR gains up to 0.2dB. Moreover, the resulting model complexity is also lower in each case. This confirms the superior performance of the proposed approach. We have also showcased this by plotting the average PSNR performance of all benchmark models against their corresponding complexity figures (in terms of FLOPs), as shown in \autoref{fig:curves}, which confirms the excellent complexity-performance trade-off achieved by the proposed framework.

\begin{figure}
    \centering
\footnotesize
    
		\begin{minipage}{0.485\linewidth}
		  \centering
		  \includegraphics[width=0.9\linewidth]{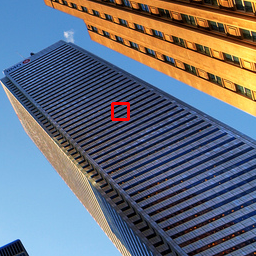}
            img\_012 from Urban100 \cite{huang2015single}
	  \end{minipage}
      \begin{minipage}{0.485\linewidth}
		  \centering
		  \includegraphics[width=0.9\linewidth]{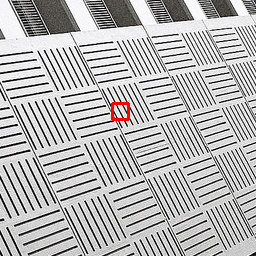}
            img\_092 from Urban100 \cite{huang2015single}
	  \end{minipage}

 	\begin{minipage}{0.485\linewidth}
		  \centering
            
		  \includegraphics[width=0.9\linewidth]{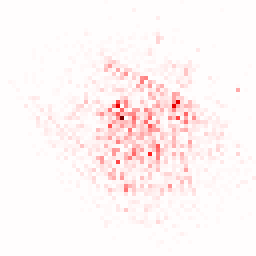}
            SRFormer\cite{zhou2023srformer}
	  \end{minipage}
 	\begin{minipage}{0.485\linewidth}
		  \centering
		  \includegraphics[width=0.9\linewidth]{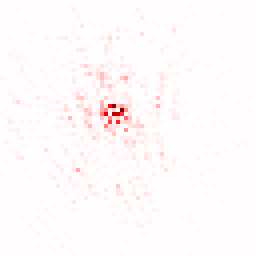}
            SRFormer\cite{zhou2023srformer}
	  \end{minipage}

              \begin{minipage}{0.485\linewidth}
		  \centering
            
	      \includegraphics[width=0.9\linewidth]{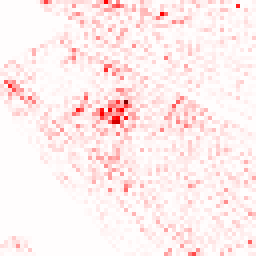}
            SRFormer-HiT\cite{zhang2024hit}
	  \end{minipage}
      \begin{minipage}{0.485\linewidth}
		  \centering
            \includegraphics[width=0.9\linewidth]{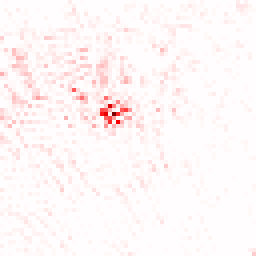}
            SRFormer-HiT\cite{zhang2024hit}
	  \end{minipage}

  	\begin{minipage}{0.485\linewidth}
		  \centering
            
	      \includegraphics[width=0.9\linewidth]{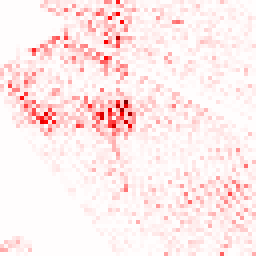}
            SRFormer-C2D (ours)
	  \end{minipage}
      \begin{minipage}{0.485\linewidth}
		  \centering
	      \includegraphics[width=0.9\linewidth]{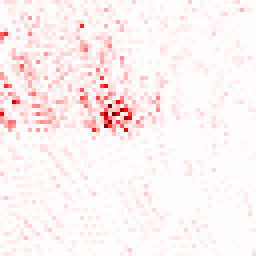}
            SRFormer-C2D (ours)
	  \end{minipage}

    \caption{The illustration of the local attribution maps (LAM) comparisons by using the Local Attribution Maps tool\cite{gu2021interpreting}.}
    \label{fig:lam}
\end{figure}

\vspace{5pt}
\noindent\textbf{Qualitative results.} Visual comparisons between the generated results by C2D-ISR optimized, HiT optimized and the lightweight backbone models are provided in \autoref{fig:Visualcomparisons}. It can be observed that C2D-ISR offers better reconstruction results compared to the benchmark methods, with fewer blocky or structural artifacts. We further utilize Local Attribution Maps (LAM) \cite{gu2021interpreting} to evaluate the effectiveness of information aggregation. Specifically, as illustrated in \autoref{fig:lam}, we apply LAM to SRFormer \cite{zhou2023srformer}, SRFormer-HiT \cite{zhang2024hit}, and SRFormer-C2D for the same target regions highlighted by red boxes. The visualization results, where larger informative areas indicate stronger aggregation capability, demonstrate that our method captures a broader range of information, which can potentially enhance SR performance.

\begin{table}[!t]
\centering
\resizebox{\linewidth}{!}{
\begin{tabular}{l|rr|rr}
\toprule

  Method ($\times 4$) & \#Para. (K) &FLOPs (G) & PSNR (dB) &SSIM  \\ \midrule
 \midrule
 v1.1-w/o C & 0&0 &  \textcolor{red}{-0.13}& \textcolor{red}{-0.0038} \\ 
 v1.2-w/o D & \textcolor{red}{+188}&\textcolor{red}{+143.3} & \textcolor{blue}{+0.05}&\textcolor{blue}{+0.0010}  \\ \midrule
 v2.1-w/o HiE & \textcolor{blue}{-2}&\textcolor{blue}{-0.3} & \textcolor{red}{-0.04}&\textcolor{red}{-0.0011}  \\ 
 v2.2-w/o CSC & \textcolor{red}{+46}&\textcolor{red}{+9.2} & \textcolor{blue}{+0.02}&\textcolor{blue}{+0.0002}  \\ \midrule
 v3.1-w/o U-Net & \textcolor{blue}{-29}&\textcolor{blue}{-3.4} &  \textcolor{red}{-0.06}&\textcolor{red}{-0.0011} \\ 
 v3.2-wo1 & 0&\textcolor{blue}{-0.7} & \textcolor{red}{-0.11}&\textcolor{red}{-0.0033} \\ 
 v3.3-wo2 & 0&0 &  \textcolor{red}{-0.05}&\textcolor{red}{-0.0014} \\ 
 v3.4-wo3 & 0&\textcolor{red}{+1.6} &  \textcolor{red}{-0.03}&\textcolor{red}{-0.0012} \\  \midrule
 \textbf{(ours)} & \textbf{849}&\textbf{57.0} & \textbf{26.96}&\textbf{0.8104}\\ \bottomrule
\end{tabular}}
\caption{Ablation study results on the Urban100 \cite{huang2015single} dataset. Here we report the PSNR/SSIM and complexity figure differences between each test variant and the original C2D-ISR model. All results are based on the $\times 4$ task and the SRFormer backbone only.}

\label{tbl:ablationstudy}

\end{table}

\subsection{Ablation Study}

We have designed an ablation experiment to verify three main contributions of this work.

\vspace{5pt}

\noindent{\textbf{C2D training strategy.}}
To test the proposed C2D training strategy, we created the following variants: (v1.1-w/o C) without the pre-training on continuous scales; (v1.2-w/o D) without fine-tuning on the discrete scales, i.e., using HIIF-L for all ISR tasks. Based on the results in \autoref{tbl:ablationstudy}, we can confirm that learning from continuous scales plays a crucial role in improving single-scale performance, with v1.1 significantly outperformed by the full C2D model. Moreover, the continuous-scale upsampler (v1.2) exhibits much higher computational complexity, requiring approximately four times the FLOPs, compared to full C2D. 


\vspace{5pt}
\noindent{\textbf{HiET layer.}}
To verify the effect of our layer design, we modified the HiET layer to obtain: (v2.1-w/o HiE) by removing the hierarchical encoding embedded with the input feature and (v2.2-w/o CSC) by replacing the CSC module with the SCC proposed in \cite{zhang2024hit}. From \autoref{tbl:ablationstudy}, we can observe that embedding hierarchical encoding (based on v2.1) does enhance reconstruction performance while introducing only minimal parameter overhead. Although utilizing the full SCC structure (v2.2) provides a minor performance improvement, it evidently increases the number of parameters and the FLOPs - this shows the better performance-complexity trade-off of the HiET layer structure.

\vspace{5pt}

\noindent{\textbf{HiET Block.}} To evaluate the contribution of our block design, we replaced the new U-Net architecture with the original linear structure in each backbone model (v3.1-w/o U-Net). We also adjusted the window size order and obtained (v3.2-wo1) [32, 16, 8, 8, 16, 32], (v3.3-wo2) [8, 32, 64, 64, 32, 8] and (v3.4-wo3) [64, 64, 64, 64, 64, 64]. Based on the results in \autoref{tbl:ablationstudy}, we can conclude that all these variants are outperformed by our design and parameter selection.

\section{Conclusion}

In this work, we present a novel optimization framework, C2D-ISR, designed to enhance lightweight attention-based ISR models. By leveraging continuous-scale pre-training, C2D-ISR enables the model to learn richer inter-scale correlations, leading to improved generalization and feature representation. 
To further boost SR performance, we designed a new Transformer architecture, HiET, which integrates hierarchical encoding mechanisms to capture multi-scale information more effectively than previous attention-based approaches. We have also employed a modified U-Net structure enabling adaptive window sizes at different levels, which achieves better multi-scale feature fusion with minimal additional computational overhead. By evaluating the proposed framework on three popular attention-based backbones, SwinIR-L, SRFormer-L, and MambaIRv2-L, we have confirmed the superior performance of our approach. C2D-ISR consistently outperforms the SOTA optimization framework, HiT, and the corresponding unoptimized lightweight SR models, with PSNR gains of up to 0.2dB (over HiT). Future work should focus on further improving computational efficiency, optimizing the efficiency of FFN, and extending the proposed framework to video super-resolution.

\section*{Acknowledgments}
The authors appreciate the funding from Netflix Inc., the University of Bristol, and the UKRI MyWorld Strength in Places Programme (SIPF00006/1).

{
    \small
    \bibliographystyle{ieeenat_fullname}
    \bibliography{main}

\begin{thebibliography}{66}
\providecommand{\natexlab}[1]{#1}
\providecommand{\url}[1]{\texttt{#1}}
\expandafter\ifx\csname urlstyle\endcsname\relax
  \providecommand{\doi}[1]{doi: #1}\else
  \providecommand{\doi}{doi: \begingroup \urlstyle{rm}\Url}\fi

\bibitem[Afonso et~al.(2018)Afonso, Zhang, and Bull]{afonso2018video}
Mariana Afonso, Fan Zhang, and David~R Bull.
\newblock Video compression based on spatio-temporal resolution adaptation.
\newblock \emph{IEEE Transactions on Circuits and Systems for Video Technology}, 29\penalty0 (1):\penalty0 275--280, 2018.

\bibitem[Agustsson and Timofte(2017)]{agustsson2017ntire}
Eirikur Agustsson and Radu Timofte.
\newblock {NTIRE} 2017 challenge on single image super-resolution: Dataset and study.
\newblock In \emph{Proceedings of the IEEE conference on computer vision and pattern recognition workshops}, pages 126--135, 2017.

\bibitem[Ahn et~al.(2018)Ahn, Kang, and Sohn]{ahn2018fast}
Namhyuk Ahn, Byungkon Kang, and Kyung-Ah Sohn.
\newblock Fast, accurate, and lightweight super-resolution with cascading residual network.
\newblock In \emph{Proceedings of the European conference on computer vision (ECCV)}, pages 252--268, 2018.

\bibitem[Bevilacqua et~al.(2012)Bevilacqua, Roumy, Guillemot, and Morel]{bevilacqua2012low}
Marco Bevilacqua, Aline Roumy, Christine Guillemot, and Marie-Line~Alberi Morel.
\newblock Low-complexity single-image super-resolution based on nonnegative neighbor embedding.
\newblock In \emph{British Machine Vision Conference (BMVC)}, 2012.

\bibitem[Chen et~al.(2023)Chen, Wang, Zhou, Qiao, and Dong]{chen2023hat}
Xiangyu Chen, Xintao Wang, Jiantao Zhou, Yu Qiao, and Chao Dong.
\newblock Activating more pixels in image super-resolution transformer.
\newblock In \emph{Proceedings of the IEEE/CVF Conference on Computer Vision and Pattern Recognition (CVPR)}, pages 22367--22377, 2023.

\bibitem[Chen et~al.(2021)Chen, Liu, and Wang]{chen2021learning}
Yinbo Chen, Sifei Liu, and Xiaolong Wang.
\newblock Learning continuous image representation with local implicit image function.
\newblock In \emph{Proceedings of the IEEE/CVF conference on computer vision and pattern recognition}, pages 8628--8638, 2021.

\bibitem[Choi et~al.(2023)Choi, Lee, and Yang]{choi2023n}
Haram Choi, Jeongmin Lee, and Jihoon Yang.
\newblock N-gram in swin transformers for efficient lightweight image super-resolution.
\newblock In \emph{Proceedings of the IEEE/CVF conference on computer vision and pattern recognition}, pages 2071--2081, 2023.

\bibitem[Conde et~al.(2022)Conde, Choi, Burchi, and Timofte]{conde2022swin2sr}
Marcos~V Conde, Ui-Jin Choi, Maxime Burchi, and Radu Timofte.
\newblock {Swin2SR}: Swinv2 transformer for compressed image super-resolution and restoration.
\newblock In \emph{European Conference on Computer Vision}, pages 669--687. Springer, 2022.

\bibitem[Dong et~al.(2015)Dong, Loy, He, and Tang]{dong2015image}
Chao Dong, Chen~Change Loy, Kaiming He, and Xiaoou Tang.
\newblock Image super-resolution using deep convolutional networks.
\newblock \emph{IEEE transactions on pattern analysis and machine intelligence}, 38\penalty0 (2):\penalty0 295--307, 2015.

\bibitem[Du et~al.(2022)Du, Liu, Liu, Tang, Wu, and Fu]{du2022fast}
Zongcai Du, Ding Liu, Jie Liu, Jie Tang, Gangshan Wu, and Lean Fu.
\newblock Fast and memory-efficient network towards efficient image super-resolution.
\newblock In \emph{Proceedings of the IEEE/CVF Conference on Computer Vision and Pattern Recognition}, pages 853--862, 2022.

\bibitem[Fang et~al.(2022{\natexlab{a}})Fang, Hu, and Hu]{fang2022cross}
Hangxiang Fang, Xinyi Hu, and Haoji Hu.
\newblock Cross knowledge distillation for image super-resolution.
\newblock In \emph{Proceedings of the 2022 6th International Conference on Video and Image Processing}, pages 162--168, 2022{\natexlab{a}}.

\bibitem[Fang et~al.(2022{\natexlab{b}})Fang, Lin, Chen, and Zeng]{fang2022hybrid}
Jinsheng Fang, Hanjiang Lin, Xinyu Chen, and Kun Zeng.
\newblock A hybrid network of cnn and transformer for lightweight image super-resolution.
\newblock In \emph{Proceedings of the IEEE/CVF conference on computer vision and pattern recognition}, pages 1103--1112, 2022{\natexlab{b}}.

\bibitem[Georgescu et~al.(2023)Georgescu, Ionescu, Miron, Savencu, Ristea, Verga, and Khan]{georgescu2023multimodal}
Mariana-Iuliana Georgescu, Radu~Tudor Ionescu, Andreea-Iuliana Miron, Olivian Savencu, Nicolae-C{\u{a}}t{\u{a}}lin Ristea, Nicolae Verga, and Fahad~Shahbaz Khan.
\newblock Multimodal multi-head convolutional attention with various kernel sizes for medical image super-resolution.
\newblock In \emph{Proceedings of the IEEE/CVF winter conference on applications of computer vision}, pages 2195--2205, 2023.

\bibitem[Gu and Dong(2021)]{gu2021interpreting}
Jinjin Gu and Chao Dong.
\newblock Interpreting super-resolution networks with local attribution maps.
\newblock In \emph{Proceedings of the IEEE/CVF Conference on Computer Vision and Pattern Recognition}, pages 9199--9208, 2021.

\bibitem[Guo et~al.(2024{\natexlab{a}})Guo, Guo, Zha, Zhang, Li, Dai, Xia, and Li]{guo2024mambairv2}
Hang Guo, Yong Guo, Yaohua Zha, Yulun Zhang, Wenbo Li, Tao Dai, Shu-Tao Xia, and Yawei Li.
\newblock Mambairv2: Attentive state space restoration.
\newblock \emph{arXiv preprint arXiv:2411.15269}, 2024{\natexlab{a}}.

\bibitem[Guo et~al.(2024{\natexlab{b}})Guo, Li, Dai, Ouyang, Ren, and Xia]{guo2024mambair}
Hang Guo, Jinmin Li, Tao Dai, Zhihao Ouyang, Xudong Ren, and Shu-Tao Xia.
\newblock Mambair: A simple baseline for image restoration with state-space model.
\newblock In \emph{ECCV}, 2024{\natexlab{b}}.

\bibitem[He et~al.(2020)He, Dai, Lu, Jiang, and Xia]{he2020fakd}
Zibin He, Tao Dai, Jian Lu, Yong Jiang, and Shu-Tao Xia.
\newblock {FAKD}: Feature-affinity based knowledge distillation for efficient image super-resolution.
\newblock In \emph{2020 IEEE International Conference on Image Processing (ICIP)}, pages 518--522. IEEE, 2020.

\bibitem[Hu et~al.(2019)Hu, Mu, Zhang, Wang, Tan, and Sun]{hu2019meta}
Xuecai Hu, Haoyuan Mu, Xiangyu Zhang, Zilei Wang, Tieniu Tan, and Jian Sun.
\newblock {Meta-SR}: A magnification-arbitrary network for super-resolution.
\newblock In \emph{Proceedings of the IEEE/CVF conference on computer vision and pattern recognition}, pages 1575--1584, 2019.

\bibitem[Huang et~al.(2015)Huang, Singh, and Ahuja]{huang2015single}
Jia-Bin Huang, Abhishek Singh, and Narendra Ahuja.
\newblock Single image super-resolution from transformed self-exemplars.
\newblock In \emph{Proceedings of the IEEE conference on computer vision and pattern recognition}, pages 5197--5206, 2015.

\bibitem[Hui et~al.(2019)Hui, Gao, Yang, and Wang]{hui2019lightweight}
Zheng Hui, Xinbo Gao, Yunchu Yang, and Xiumei Wang.
\newblock Lightweight image super-resolution with information multi-distillation network.
\newblock In \emph{Proceedings of the 27th acm international conference on multimedia}, pages 2024--2032, 2019.

\bibitem[Jiang et~al.(2024)Jiang, Nawa{\l}a, Zhang, and Bull]{jiang2023compressing}
Yuxuan Jiang, Jakub Nawa{\l}a, Fan Zhang, and David Bull.
\newblock Compressing deep image super-resolution models.
\newblock In \emph{2024 Picture Coding Symposium (PCS)}, pages 1--5. IEEE, 2024.

\bibitem[Jiang et~al.(2025{\natexlab{a}})Jiang, Feng, Zhang, and Bull]{jiang2024mtkd}
Yuxuan Jiang, Chen Feng, Fan Zhang, and David Bull.
\newblock {MTKD}: Multi-teacher knowledge distillation for image super-resolution.
\newblock In \emph{{ECCV}}, pages 364--382. Springer, 2025{\natexlab{a}}.

\bibitem[Jiang et~al.(2025{\natexlab{b}})Jiang, Kwan, Peng, Gao, Zhang, Zhu, Sole, and Bull]{jiang2024hiif}
Yuxuan Jiang, Ho~Man Kwan, Tianhao Peng, Ge Gao, Fan Zhang, Xiaoqing Zhu, Joel Sole, and David Bull.
\newblock Hiif: Hierarchical encoding based implicit image function for continuous super-resolution.
\newblock \emph{Proceedings of the IEEE/CVF Conference on Computer Vision and Pattern Recognition}, 2025{\natexlab{b}}.

\bibitem[Kim et~al.(2016{\natexlab{a}})Kim, Lee, and Lee]{kim2016accurate}
Jiwon Kim, Jung~Kwon Lee, and Kyoung~Mu Lee.
\newblock Accurate image super-resolution using very deep convolutional networks.
\newblock In \emph{Proceedings of the IEEE conference on computer vision and pattern recognition}, pages 1646--1654, 2016{\natexlab{a}}.

\bibitem[Kim et~al.(2016{\natexlab{b}})Kim, Lee, and Lee]{kim2016deeply}
Jiwon Kim, Jung~Kwon Lee, and Kyoung~Mu Lee.
\newblock Deeply-recursive convolutional network for image super-resolution.
\newblock In \emph{Proceedings of the IEEE conference on computer vision and pattern recognition}, pages 1637--1645, 2016{\natexlab{b}}.

\bibitem[Kingma and Ba(2014)]{kingma2014adam}
Diederik~P Kingma and Jimmy Ba.
\newblock Adam: a method for stochastic optimization.
\newblock \emph{arXiv preprint arXiv:1412.6980}, 2014.

\bibitem[Kwan et~al.(2023)Kwan, Gao, Zhang, Gower, and Bull]{kwan2023hinerv}
Ho~Man Kwan, Ge Gao, Fan Zhang, Andrew Gower, and David Bull.
\newblock {HiNeRV}: Video compression with hierarchical encoding based neural representation.
\newblock In \emph{NeurIPS}, 2023.

\bibitem[Kwan et~al.(2024)Kwan, Gao, Zhang, Gower, and Bull]{kwan2024nvrc}
Ho~Man Kwan, Ge Gao, Fan Zhang, Andrew Gower, and David Bull.
\newblock {NVRC}: Neural video representation compression.
\newblock \emph{arXiv preprint arXiv:2409.07414}, 2024.

\bibitem[Lai et~al.(2017)Lai, Huang, Ahuja, and Yang]{lai2017deep}
Wei-Sheng Lai, Jia-Bin Huang, Narendra Ahuja, and Ming-Hsuan Yang.
\newblock Deep laplacian pyramid networks for fast and accurate super-resolution.
\newblock In \emph{Proceedings of the IEEE conference on computer vision and pattern recognition}, pages 624--632, 2017.

\bibitem[Lee and Jin(2022)]{lee2022local}
Jaewon Lee and Kyong~Hwan Jin.
\newblock Local texture estimator for implicit representation function.
\newblock In \emph{Proceedings of the IEEE/CVF conference on computer vision and pattern recognition}, pages 1929--1938, 2022.

\bibitem[Li et~al.(2024)Li, Luo, Xu, and Wong]{li2024asmr}
Jason Chun~Lok Li, Steven Tin~Sui Luo, Le Xu, and Ngai Wong.
\newblock {ASMR:} activation-sharing multi-resolution coordinate networks for efficient inference.
\newblock In \emph{{ICLR}}. OpenReview.net, 2024.

\bibitem[Li et~al.(2021)Li, Sixou, and Peyrin]{li2021review}
Yufei Li, Bruno Sixou, and Francois Peyrin.
\newblock A review of the deep learning methods for medical images super resolution problems.
\newblock \emph{Irbm}, 42\penalty0 (2):\penalty0 120--133, 2021.

\bibitem[Liang et~al.(2021)Liang, Cao, Sun, Zhang, Van~Gool, and Timofte]{liang2021swinir}
Jingyun Liang, Jiezhang Cao, Guolei Sun, Kai Zhang, Luc Van~Gool, and Radu Timofte.
\newblock {SwinIR}: Image restoration using swin transformer.
\newblock In \emph{Proceedings of the IEEE/CVF international conference on computer vision}, pages 1833--1844, 2021.

\bibitem[Liang et~al.(2022)Liang, Cao, Fan, Zhang, Ranjan, Li, Timofte, and Van~Gool]{liang2022vrt}
Jingyun Liang, Jiezhang Cao, Yuchen Fan, Kai Zhang, Rakesh Ranjan, Yawei Li, Radu Timofte, and Luc Van~Gool.
\newblock {VRT}: A video restoration transformer.
\newblock \emph{arXiv preprint arXiv:2201.12288}, 2022.

\bibitem[Lim et~al.(2017)Lim, Son, Kim, Nah, and Mu~Lee]{lim2017enhanced}
Bee Lim, Sanghyun Son, Heewon Kim, Seungjun Nah, and Kyoung Mu~Lee.
\newblock Enhanced deep residual networks for single image super-resolution.
\newblock In \emph{Proceedings of the IEEE conference on computer vision and pattern recognition workshops}, pages 136--144, 2017.

\bibitem[Liu et~al.(2020)Liu, Tang, and Wu]{liu2020residual}
Jie Liu, Jie Tang, and Gangshan Wu.
\newblock Residual feature distillation network for lightweight image super-resolution.
\newblock In \emph{Computer vision--ECCV 2020 workshops: Glasgow, UK, August 23--28, 2020, proceedings, part III 16}, pages 41--55. Springer, 2020.

\bibitem[Loshchilov and Hutter(2016)]{loshchilov2016sgdr}
Ilya Loshchilov and Frank Hutter.
\newblock Sgdr: Stochastic gradient descent with warm restarts.
\newblock \emph{arXiv preprint arXiv:1608.03983}, 2016.

\bibitem[Lu et~al.(2022)Lu, Li, Liu, Huang, Zhang, and Zeng]{lu2022transformer}
Zhisheng Lu, Juncheng Li, Hong Liu, Chaoyan Huang, Linlin Zhang, and Tieyong Zeng.
\newblock Transformer for single image super-resolution.
\newblock In \emph{Proceedings of the IEEE/CVF conference on computer vision and pattern recognition}, pages 457--466, 2022.

\bibitem[Luo et~al.(2020)Luo, Xie, Zhang, Qu, Li, and Fu]{luo2020latticenet}
Xiaotong Luo, Yuan Xie, Yulun Zhang, Yanyun Qu, Cuihua Li, and Yun Fu.
\newblock {LatticeNet}: Towards lightweight image super-resolution with lattice block.
\newblock In \emph{Computer Vision--ECCV 2020: 16th European Conference, Glasgow, UK, August 23--28, 2020, Proceedings, Part XXII 16}, pages 272--289. Springer, 2020.

\bibitem[Martin et~al.(2001)Martin, Fowlkes, Tal, and Malik]{martin2001database}
David Martin, Charless Fowlkes, Doron Tal, and Jitendra Malik.
\newblock A database of human segmented natural images and its application to evaluating segmentation algorithms and measuring ecological statistics.
\newblock In \emph{Proceedings Eighth IEEE International Conference on Computer Vision. ICCV 2001}, pages 416--423. IEEE, 2001.

\bibitem[Matsui et~al.(2017)Matsui, Ito, Aramaki, Fujimoto, Ogawa, Yamasaki, and Aizawa]{matsui2017sketch}
Yusuke Matsui, Kota Ito, Yuji Aramaki, Azuma Fujimoto, Toru Ogawa, Toshihiko Yamasaki, and Kiyoharu Aizawa.
\newblock Sketch-based manga retrieval using manga109 dataset.
\newblock \emph{Multimedia tools and applications}, 76:\penalty0 21811--21838, 2017.

\bibitem[Park et~al.(2025)Park, Soh, and Cho]{park2025efficient}
Karam Park, Jae~Woong Soh, and Nam~Ik Cho.
\newblock Efficient attention-sharing information distillation transformer for lightweight single image super-resolution.
\newblock \emph{arXiv preprint arXiv:2501.15774}, 2025.

\bibitem[Paszke et~al.(2019)Paszke, Gross, Massa, Lerer, Bradbury, Chanan, Killeen, Lin, Gimelshein, Antiga, et~al.]{paszke2019pytorch}
Adam Paszke, Sam Gross, Francisco Massa, Adam Lerer, James Bradbury, Gregory Chanan, Trevor Killeen, Zeming Lin, Natalia Gimelshein, Luca Antiga, et~al.
\newblock Pytorch: An imperative style, high-performance deep learning library.
\newblock \emph{Advances in neural information processing systems}, 32, 2019.

\bibitem[Ren et~al.(2024)Ren, Li, Guo, Li, Zhao, and Chen]{ren2024mambacsr}
Yulin Ren, Xin Li, Mengxi Guo, Bingchen Li, Shijie Zhao, and Zhibo Chen.
\newblock {MambaCSR}: Dual-interleaved scanning for compressed image super-resolution with ssms, 2024.

\bibitem[Ronneberger et~al.(2015)Ronneberger, Fischer, and Brox]{ronneberger2015u}
Olaf Ronneberger, Philipp Fischer, and Thomas Brox.
\newblock U-net: Convolutional networks for biomedical image segmentation.
\newblock In \emph{Medical image computing and computer-assisted intervention--MICCAI 2015: 18th international conference, Munich, Germany, October 5-9, 2015, proceedings, part III 18}, pages 234--241. Springer, 2015.

\bibitem[Shi et~al.(2024)Shi, Xia, Jin, Wang, Zhao, Xia, Xiao, and Yang]{shi2024vmambair}
Yuan Shi, Bin Xia, Xiaoyu Jin, Xing Wang, Tianyu Zhao, Xin Xia, Xuefeng Xiao, and Wenming Yang.
\newblock {VmambaIR}: Visual state space model for image restoration.
\newblock \emph{arXiv preprint arXiv:2403.11423}, 2024.

\bibitem[Tai et~al.(2017{\natexlab{a}})Tai, Yang, and Liu]{tai2017image}
Ying Tai, Jian Yang, and Xiaoming Liu.
\newblock Image super-resolution via deep recursive residual network.
\newblock In \emph{Proceedings of the IEEE conference on computer vision and pattern recognition}, pages 3147--3155, 2017{\natexlab{a}}.

\bibitem[Tai et~al.(2017{\natexlab{b}})Tai, Yang, Liu, and Xu]{tai2017memnet}
Ying Tai, Jian Yang, Xiaoming Liu, and Chunyan Xu.
\newblock Memnet: A persistent memory network for image restoration.
\newblock In \emph{Proceedings of the IEEE international conference on computer vision}, pages 4539--4547, 2017{\natexlab{b}}.

\bibitem[Timofte et~al.(2017)Timofte, Agustsson, Van~Gool, Yang, and Zhang]{timofte2017ntire}
Radu Timofte, Eirikur Agustsson, Luc Van~Gool, Ming-Hsuan Yang, and Lei Zhang.
\newblock {NTIRE} 2017 challenge on single image super-resolution: Methods and results.
\newblock In \emph{Proceedings of the IEEE conference on computer vision and pattern recognition workshops}, pages 114--125, 2017.

\bibitem[Wang et~al.(2023{\natexlab{a}})Wang, Chen, Ni, Liu, and Liu]{wang2023omni}
Hang Wang, Xuanhong Chen, Bingbing Ni, Yutian Liu, and Jinfan Liu.
\newblock Omni aggregation networks for lightweight image super-resolution.
\newblock In \emph{Proceedings of the IEEE/CVF Conference on Computer Vision and Pattern Recognition}, pages 22378--22387, 2023{\natexlab{a}}.

\bibitem[Wang et~al.(2023{\natexlab{b}})Wang, Zhang, Qin, Van~Gool, and Fu]{wang2023global}
Huan Wang, Yulun Zhang, Can Qin, Luc Van~Gool, and Yun Fu.
\newblock Global aligned structured sparsity learning for efficient image super-resolution.
\newblock \emph{IEEE transactions on pattern analysis and machine intelligence}, 45\penalty0 (9):\penalty0 10974--10989, 2023{\natexlab{b}}.

\bibitem[Wang et~al.(2022{\natexlab{a}})Wang, Bayram, and Sertel]{wang2022comprehensive}
Peijuan Wang, Bulent Bayram, and Elif Sertel.
\newblock A comprehensive review on deep learning based remote sensing image super-resolution methods.
\newblock \emph{Earth-Science Reviews}, 232:\penalty0 104110, 2022{\natexlab{a}}.

\bibitem[Wang et~al.(2004)Wang, Bovik, Sheikh, and Simoncelli]{wang2004image}
Zhou Wang, Alan~C Bovik, Hamid~R Sheikh, and Eero~P Simoncelli.
\newblock {Image Quality Assessment}: from error visibility to structural similarity.
\newblock \emph{IEEE transactions on image processing}, 13\penalty0 (4):\penalty0 600--612, 2004.

\bibitem[Wang et~al.(2020)Wang, Chen, and Hoi]{wang2020deep}
Zhihao Wang, Jian Chen, and Steven~CH Hoi.
\newblock Deep learning for image super-resolution: A survey.
\newblock \emph{IEEE transactions on pattern analysis and machine intelligence}, 43\penalty0 (10):\penalty0 3365--3387, 2020.

\bibitem[Wang et~al.(2022{\natexlab{b}})Wang, Cun, Bao, Zhou, Liu, and Li]{wang2022uformer}
Zhendong Wang, Xiaodong Cun, Jianmin Bao, Wengang Zhou, Jianzhuang Liu, and Houqiang Li.
\newblock Uformer: A general u-shaped transformer for image restoration.
\newblock In \emph{Proceedings of the IEEE/CVF conference on computer vision and pattern recognition}, pages 17683--17693, 2022{\natexlab{b}}.

\bibitem[Xiao et~al.(2021)Xiao, Su, Yuan, Liu, Shen, and Zhang]{xiao2021satellite}
Yi Xiao, Xin Su, Qiangqiang Yuan, Denghong Liu, Huanfeng Shen, and Liangpei Zhang.
\newblock Satellite video super-resolution via multiscale deformable convolution alignment and temporal grouping projection.
\newblock \emph{IEEE Transactions on Geoscience and Remote Sensing}, 60:\penalty0 1--19, 2021.

\bibitem[Yu et~al.(2023)Yu, Li, Li, Jiang, Wu, Fan, and Liu]{yu2023dipnet}
Lei Yu, Xinpeng Li, Youwei Li, Ting Jiang, Qi Wu, Haoqiang Fan, and Shuaicheng Liu.
\newblock Dipnet: Efficiency distillation and iterative pruning for image super-resolution.
\newblock In \emph{Proceedings of the IEEE/CVF Conference on Computer Vision and Pattern Recognition}, pages 1692--1701, 2023.

\bibitem[Zeyde et~al.(2012)Zeyde, Elad, and Protter]{zeyde2012single}
Roman Zeyde, Michael Elad, and Matan Protter.
\newblock On single image scale-up using sparse-representations.
\newblock In \emph{Curves and Surfaces: 7th International Conference, Avignon, France, June 24-30, 2010, Revised Selected Papers 7}, pages 711--730. Springer, 2012.

\bibitem[Zhang et~al.(2021{\natexlab{a}})Zhang, Afonso, and Bull]{zhang2021vistra2}
Fan Zhang, Mariana Afonso, and David~R Bull.
\newblock {ViSTRA2}: Video coding using spatial resolution and effective bit depth adaptation.
\newblock \emph{Signal Processing: Image Communication}, 97:\penalty0 116355, 2021{\natexlab{a}}.

\bibitem[Zhang et~al.(2010)Zhang, Zhang, Shen, and Li]{zhang2010super}
Liangpei Zhang, Hongyan Zhang, Huanfeng Shen, and Pingxiang Li.
\newblock A super-resolution reconstruction algorithm for surveillance images.
\newblock \emph{Signal Processing}, 90\penalty0 (3):\penalty0 848--859, 2010.

\bibitem[Zhang et~al.(2022)Zhang, Zeng, Guo, and Zhang]{zhang2022efficient}
Xindong Zhang, Hui Zeng, Shi Guo, and Lei Zhang.
\newblock Efficient long-range attention network for image super-resolution.
\newblock In \emph{European conference on computer vision}, pages 649--667. Springer, 2022.

\bibitem[Zhang et~al.(2024)Zhang, Zhang, and Yu]{zhang2024hit}
Xiang Zhang, Yulun Zhang, and Fisher Yu.
\newblock Hit-sr: Hierarchical transformer for efficient image super-resolution.
\newblock In \emph{European Conference on Computer Vision}, pages 483--500. Springer, 2024.

\bibitem[Zhang et~al.(2018)Zhang, Li, Li, Wang, Zhong, and Fu]{zhang2018image}
Yulun Zhang, Kunpeng Li, Kai Li, Lichen Wang, Bineng Zhong, and Yun Fu.
\newblock Image super-resolution using very deep residual channel attention networks.
\newblock In \emph{Proceedings of the European conference on computer vision (ECCV)}, pages 286--301, 2018.

\bibitem[Zhang et~al.(2021{\natexlab{b}})Zhang, Wang, Qin, and Fu]{zhang2021learning}
Yulun Zhang, Huan Wang, Can Qin, and Yun Fu.
\newblock Learning efficient image super-resolution networks via structure-regularized pruning.
\newblock In \emph{International conference on learning representations}, 2021{\natexlab{b}}.

\bibitem[Zhou et~al.(2023)Zhou, Li, Guo, Bai, Cheng, and Hou]{zhou2023srformer}
Yupeng Zhou, Zhen Li, Chun-Le Guo, Song Bai, Ming-Ming Cheng, and Qibin Hou.
\newblock Srformer: Permuted self-attention for single image super-resolution.
\newblock In \emph{Proceedings of the IEEE/CVF International Conference on Computer Vision}, pages 12780--12791, 2023.

\bibitem[Zhu et~al.(2023)Zhu, Li, and Li]{zhu2023attention}
Qiang Zhu, Pengfei Li, and Qianhui Li.
\newblock Attention retractable frequency fusion transformer for image super resolution.
\newblock In \emph{Proceedings of the IEEE/CVF Conference on Computer Vision and Pattern Recognition}, pages 1756--1763, 2023.

\end{thebibliography}
}

\end{document}